\documentclass[11pt]{article}

\usepackage[pdftex]{graphicx}
\usepackage[pdftex]{color}

\usepackage[square]{natbib}
\bibpunct[:]{(}{)}{;}{a}{}{:}
\usepackage{amssymb}
\usepackage{color}

\usepackage{graphicx}

\newcommand{\eqref}[1]{Formula~ \ref{#1}}
\newcommand{\secref}[1]{\S \ref{#1}}
\newcommand{\figref}[1]{Figure~\ref{#1}}

\usepackage{color}
\iffalse

\newcommand{\add}[1]{\textcolor{red}{#1}}
\newcommand{\delete}[1]{\textcolor{blue}{#1}}
\else

\newcommand{\add}[1]{\textcolor{black}{#1}}
\newcommand{\delete}[1]{}
\fi

\begin{document}
\vspace*{0.2in}
\begin{flushleft}
{\Large
  \textbf\newline{
Do Neural Nets Learn Statistical Laws behind Natural 
Language?}
}
\newline
\\
Shuntaro Takahashi\textsuperscript{1\dag},
Kumiko Tanaka-Ishii\textsuperscript{2\ddag *}
\\
\bigskip
\textbf{1}
The University of Tokyo, Graduate School of Frontier Sciences,
Chiba 277-8563, Japan.
\\
\textbf{2}
The University of Tokyo, Research Center for Advanced Science and Technology,
Tokyo 153-8904, Japan.
\\
\bigskip
* kumiko@cl.rcast.u-tokyo.ac.jp

\end{flushleft}

\section*{Abstract}
The performance of deep learning in natural language processing has been spectacular, 
but the reasons for this success remain unclear because of the inherent complexity of 
deep learning. This paper provides empirical evidence of its effectiveness and of a 
limitation of neural networks for language engineering. Precisely, we demonstrate that a 
neural language model based on long short-term memory (LSTM) effectively 
reproduces Zipf's law and Heaps' law, two representative statistical properties 
underlying natural language. We discuss the quality of reproducibility and the 
emergence of Zipf's law and Heaps' law as training progresses. We also point out that 
the neural language model has a limitation in reproducing long-range correlation, 
another statistical property of natural language. This understanding could provide a 
direction for improving the architectures of neural networks. 

%\linenumbers

\section{Introduction}
Deep learning has performed spectacularly in various natural language processing tasks 
such as machine translation \citep{Wu_2016}, text summarization 
\citep{Rush:EMNLP:2015}, dialogue systems \citep{Serban:AAAI:2015}, and question 
answering \citep{Tan:ICLR:2015}. A fundamental question that we ask, however, is why 
deep learning is such an effective approach for natural language processing. In contrast 
to the progress made in applying deep learning, our understanding of the reasons for its 
effectiveness remains limited because of its inherent complexity.

One approach to tackling this problem is mathematical analysis of the potential of 
neural networks \citep{NIPS2014_5422, ICML:Cohen+Shashua:2016, 
cohen2016expressive, bianchini2014complexity, NIPS:Poole+etal:2016, 
Lin_Tegmark_2016,Schwab_2016}. Here, we take a different empirical approach based 
on the statistical properties of text generated by neural networks. Precisely, we compare 
the statistical properties of pseudo-text generated by a neural language model with those 
of the real text with which the model is trained.

We have found that two well acknowledged statistical laws of natural language---Zipf's 
law \citep{zipf} and Heaps' law 
\citep{heaps}\citep{Herdan1964}\citep{Guiraud1954}---almost hold for the pseudo-text 
generated by a neural language model. This finding is notable because previous 
language models, such as Markov models, cannot reproduce such properties, and 
mathematical models, which are designed to reproduce statistical laws 
\citep{pitman}\citep{simon55}, are also limited in their purpose. As compared with those 
models, neural language models are far more advanced in satisfying the statistical laws. 
We find a shortcoming of neural language models, however, in that the generated 
pseudo-text has a limitation with respect to satisfying a third statistical property, the 
long-range correlation. The analyses described in this paper contribute to our 
understanding of the performance of neural networks and provide guidance as to how 
we can improve models.

\section{Neural language models generate text following Zipf's law 
and Heaps' law}
\subsection{Neural language model}

\label{sec:stacked}

We constructed a neural language model that learns from a corpus and generates a 
pseudo-text, and then investigated whether the model produced any statistical laws of 
language. The language model estimates the probability of the next element of the 
sequence, $w_{i+1}$, given its past sequence or a subset as context:
\begin{equation}
P (w_{i+1}|w_{i-k}^{i}),
\end{equation}
where $k$ is the context length, and $w_{i}^{j}$ is the subsequence of text between 
the $i$th and $j$th elements. Bengio et al.\citep{bengio03} first proposed the concept of 
a neural language model, and this concept has been explored mainly with recurrent 
neural networks (RNNs) \citep{Krause}\citep{Chelba_2017}\citep{Sundermeyer_2012}. 
We construct a language model at the character level, which we denote as a stacked long 
short-term memory (LSTM) \citep{Hochreiter_1997} model. This model consists of 
three LSTM layers \add{with 256 units each} and a softmax \add{output} layer. We treat this stacked LSTM model as a 
representative of neural language models.

\add{In all experiments in this article, the model was trained to
  minimize the cross-entropy by using an Adam optimizer with the
  proposed hyper-parameters \citep{Kingma_2014}.  The context length
  $k$ was set to $128$. To avoid sample biases and hence increase the
  generalization performance, the dataset was shuffled during the
  training procedure: i.e. every one learning scan of the
    training data is conducted in a different shuffled order.}  This
  is a standard configuration with respect to previous research on
  neural language models
  \citep{Krause}\citep{Chelba_2017}\citep{Sundermeyer_2012}\citep{Lin_2016}.

\add{In the normal scheme of deep learning research, the model learns
  from all the samples of the training dataset once during an
  epoch. In this work, however, we redefined the scheme so that the
  model learns from 1\% of the training dataset during every
  epoch. That is, the model learns from all the samples every 100
  epochs. We adopted this definition because the evolutions of Zipf's
  law and Heaps' law are so fast that their corresponding behaviors
  are clearly present after the model has learned from all the samples
  once. Although we discuss this topic in \figref{dl_development}, we
  emphasize here that either this redefinition or some other approach
  was necessary to observe the model's development with respect to
  Zipf's law and Heaps' law.}

Generation of a pseudo-text begins with 128 characters in succession
as context, where the 128-character sequence exists in the original
text. One character to follow the context is chosen randomly according
to the probability distribution of the neural model's output. The
context is then shifted ahead by one character to include the latest
character. This procedure is repeated to produce a pseudo-text of 2
million characters, except in Fig.2, 20 million. \add{The
  following is an example of a generated pseudo-text: {\textit{``and
      you gracious inherites and what saist i should agge the
      guest.''}}}

\begin{figure}[t]
\begin{center}
\includegraphics[width=\textwidth]{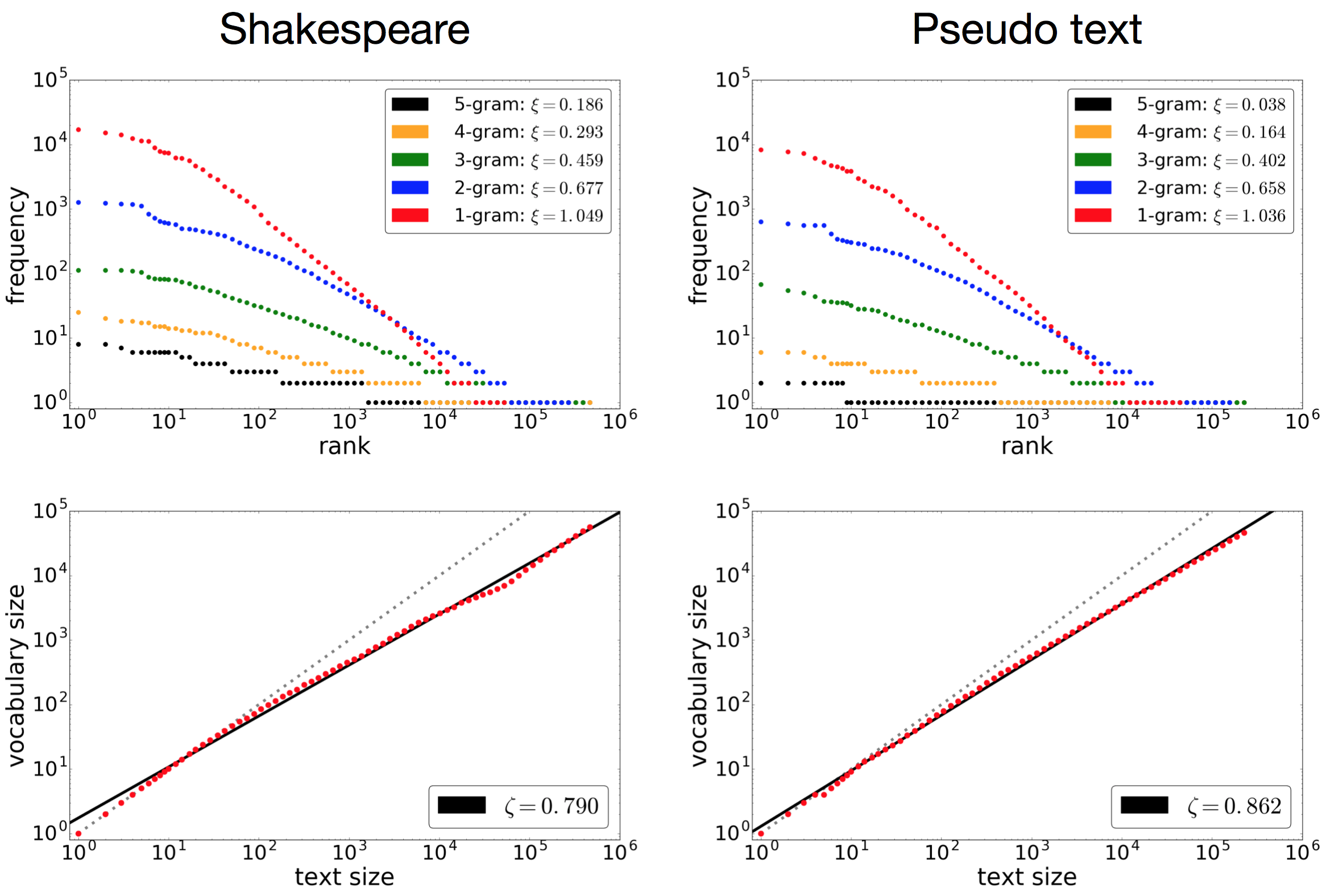}
\caption{The rank-frequency distribution and vocabulary growth of the Complete 
Works of Shakespeare (left) and the corresponding pseudo-text generated by the stacked 
LSTM model (right). All axes in this and subsequent figures in this paper are in 
logarithmic scale, and the plots were generated using logarithmic bins. The model 
learned from 4,121,423 characters in the Complete Works of Shakespeare, which was 
preprocessed as described in the main text.
  The colored sets of the plots of the figures in the first row show the rank-frequency distributions of 1,2,3,4,5 grams.
  The figures in the second row show the vocabulary growth in red.
\add{The exponents $\xi$ and $\zeta$ (defined in formulas 
\eqref{eq:zipf} and \eqref{eq:heaps}, respectively) were estimated by linear regression 
from in log-log scale.}
For all graphs, the corresponding estimated exponents are indicated in the caption, and the black solid line
in each vocabulary growth figure shows the fitted line.
The dashed line indicates a reference with an exponent of 1.0. The same applies to all other rank-frequency 
distribution and vocabulary growth plots in this paper.}
\label{shakespeare:zipfheaps}
\end{center}
\end{figure}

We chose a character-level language model because word-level models have the critical 
problem of being unable to introduce new words during generation: by definition, they 
do not generate new words unless special architectures are added. A word-level model 
typically processes all words with rarity above a certain threshold by transforming each 
into a singular symbol ``unk''. With such a model, there is a definite vocabulary size 
limit, thus destroying the tail of the rank-frequency distribution. Zipf's law and Heaps' 
law therefore cannot be reproduced with such a model. There have been discussions and 
proposals regarding this ``unk'' problem \citep{acl14} \citep{Luong_2016}, but there is 
no de facto standard approach, and the problem is not straightforward to solve. 
Therefore, we chose a character-level language model.

\add{Note that the English datasets, consisting of the Complete Works of Shakespeare 
and The Wall Street Journal (WSJ), were preprocessed according to \citep{Krause} by 
making all alphabetical characters lower case and removing all non-alphabetical 
characters except spaces. Consecutive spaces were also reduced to one space.} 

\subsection{Zipf's law and Heaps' law for pseudo-texts generated by neural  
language models}
Zipf's law and Heaps' law are two representative statistical properties of natural 
language. Zipf's law states that, given word rank $u$ and frequency $F(u)$ for a word 
of rank $u$, the following proportionality holds:
\begin{equation}
F (u) \propto u^{-\xi}.  \ \ \label{eq:zipf}
\end{equation}
This exponent $\xi$ is approximately 1.0, according to Zipf, for individual word 
occurrence (uni-grams), but, as will be shown, a power law with smaller $\xi$ values 
holds for longer word sequences (i.e., $n$-grams, including 2-grams, 3-grams, and so 
on).

Heaps' law, another statistical law of natural language, underlies the growth rate of 
vocabulary size (the number of types) with respect to text length (the number of tokens). 
Given vocabulary size $V(m)$ for a text of length $m$, Heaps' law indicates that
\begin{equation}
  V (m) \propto m^{\zeta}.   \ \ \label{eq:heaps}
\end{equation}
The power law underlying vocabulary growth was reported even before Heaps' paper 
\citep{heaps}, as in \citep{Herdan1964}\citep{Guiraud1954}, but in this paper we refer to 
the law as Heaps' law. Zipf's law and Heaps law are known to have a theoretical 
relationship, as discussed in 
\citep{BaezaYates_2000}\citep{Leijenhorst_2005}\citep{lu2010}.

The upper-left graph in \figref{shakespeare:zipfheaps} shows the rank-frequency 
distribution of the Complete Works of Shakespeare, \add{consisting of 4,121,423 
characters}, for $n$-grams ranging from uni-grams to 5-grams. As Zipf stated, the 
uni-gram distribution approximately follows a power law with an exponent of 1.0. The 
higher $n$-gram distributions also follow power laws but with smaller exponents. Note 
that intersection of the uni-gram and 2-gram distributions in the tail is typically 
observed for natural language. The lower-left graph in \figref{shakespeare:zipfheaps} 
shows the vocabulary growth of the Complete Works of Shakespeare. The red points show the 
vocabulary size $V(m)$ for every text length $m$, and the exponent $\zeta$ was 
estimated as 0.773, as shown by the black fitting line. This exponent is larger than that 
reported in previous works, and this was due to the preprocessing, as previously 
mentioned.

The graphs on the right side of \figref{shakespeare:zipfheaps} show the corresponding 
rank-frequency distribution and vocabulary growth of the pseudo-text generated by the 
stacked LSTM. The rank-frequency distribution is almost identical to that of the 
Complete Works of Shakespeare for uni-grams and 2-grams, reproducing the original shape of the 
distribution. The distributions for longer $n$-grams are also well reproduced. As for the 
vocabulary growth, the language model introduces new words according to a power law 
with a slightly larger exponent than that of the original text. This suggests a limitation 
on the recognition of words and the organization of $n$-gram sequences. These results 
indicate that the stacked LSTM can reproduce an $n$-gram structure closely resembling 
the original structure.

\begin{figure}[t]
\begin{center}
\includegraphics[width=\textwidth]{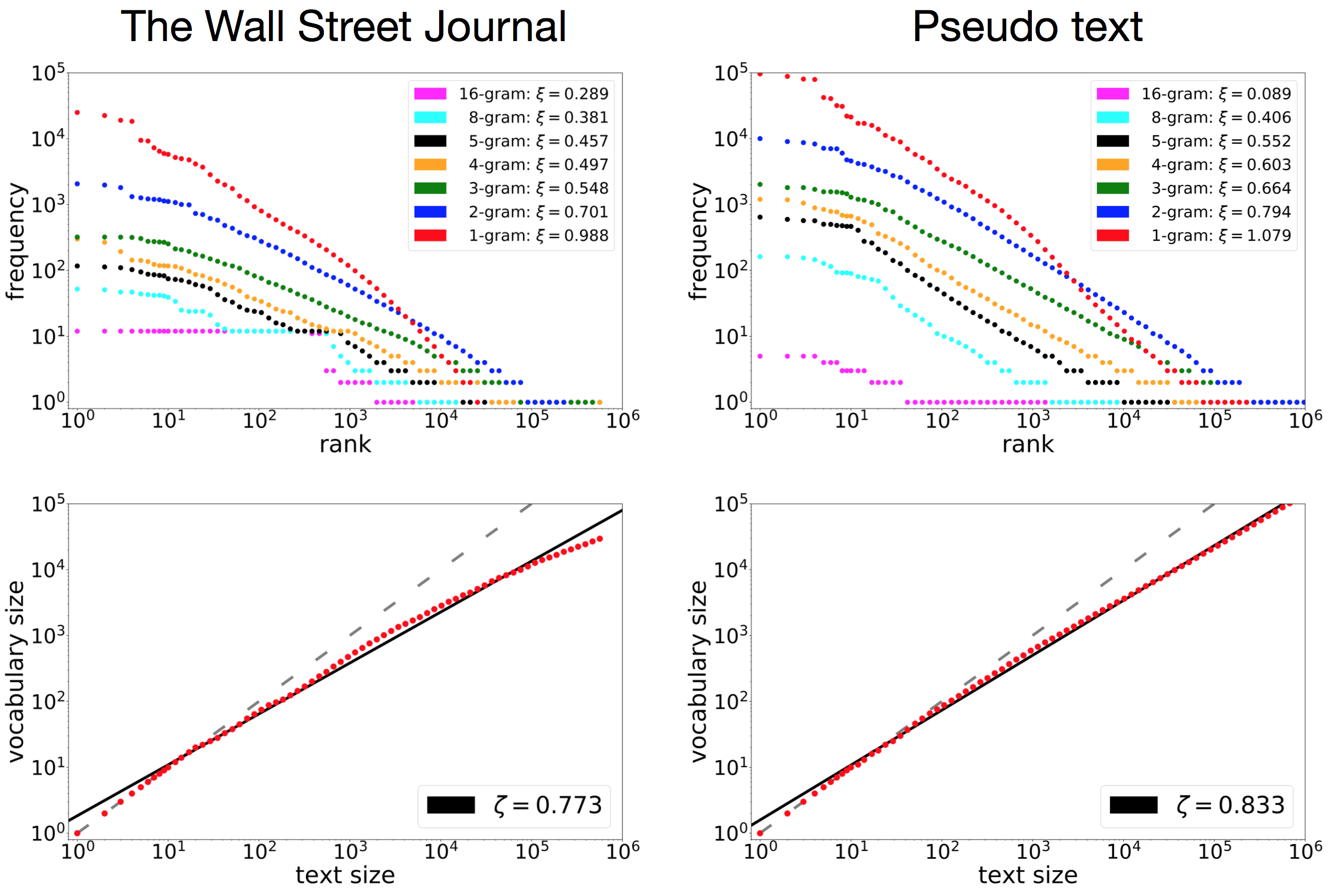}
\caption{The rank-frequency distribution and vocabulary growth of The Wall Street 
Journal (left) and the corresponding pseudo-text generated by the stacked LSTM model 
(right). The model learned from 4,780,916 characters.
The length of the pseudo-text is 20 million characters.
The rank-frequency distributions are shown for 1,2,3,4,5,8,16-grams.
The preprocessing procedure was the 
same as for the Complete Works of Shakespeare.}
\label{PTB:zipfheaps}
\end{center}
\end{figure}

\begin{figure}[t]
\begin{center}
\includegraphics[width=\textwidth]{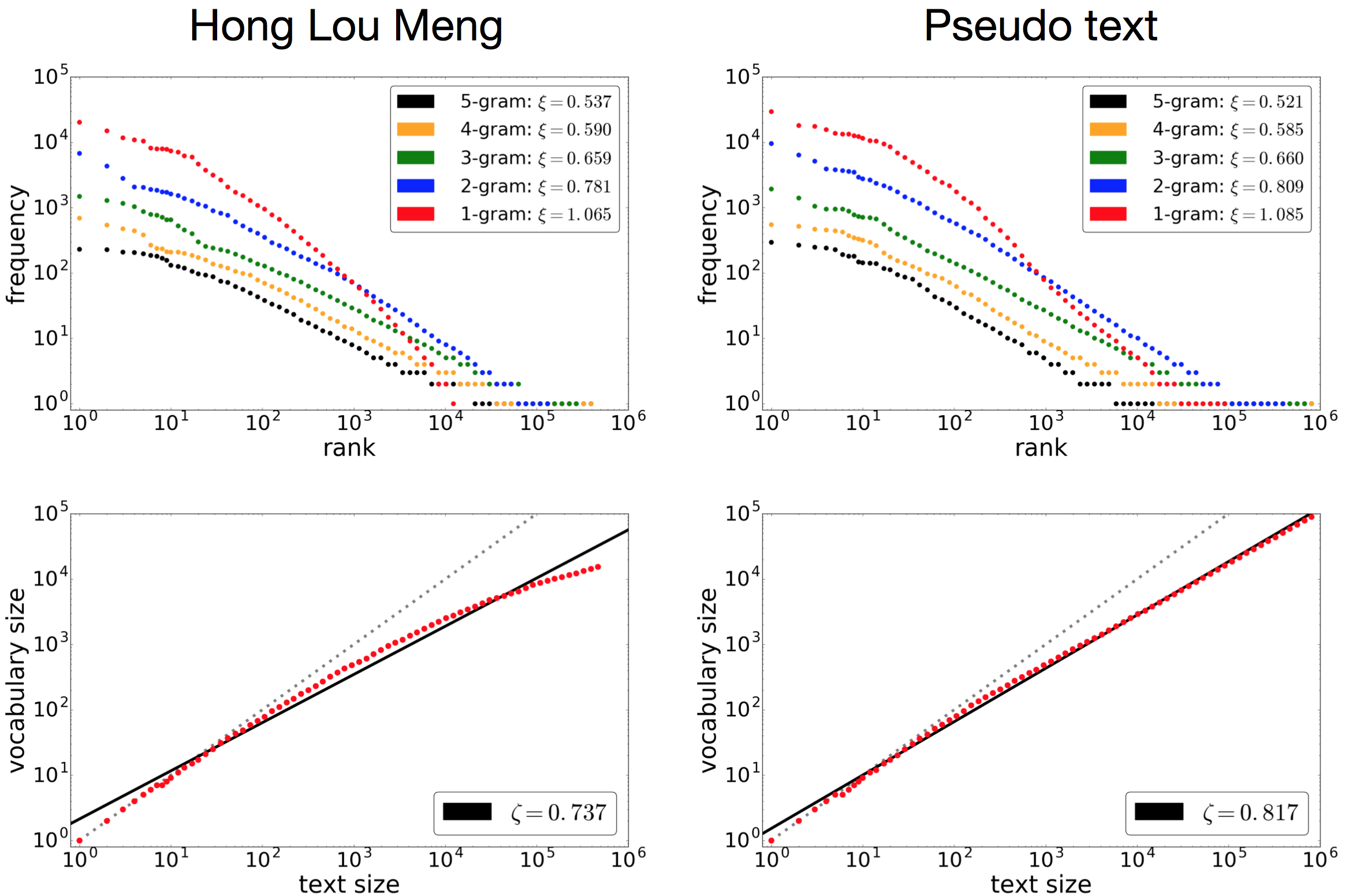}
\caption{The rank-frequency distribution and vocabulary growth of the Chinese literary 
work Hong Lou Meng, consisting of $2,932,451$ bytes (left), and the pseudo-text 
generated by the stacked LSTM (right). The text was processed at the byte level with 
word borders.}
\label{Chinese:zipfheaps}
\end{center}
\end{figure}

The potential of the stacked LSTM is still apparent even when we change the kind of 
text. \figref{PTB:zipfheaps} and \figref{Chinese:zipfheaps} show results obtained 
using The Wall Street Journal (from the Penn Tree Bank Dataset) and a Chinese 
literary text, Hong Lou Meng by X. C. Xueqin, respectively, and the corresponding 
pseudo-texts generated by the stacked LSTM. The WSJ text of 
$4,780,916$ characters was subjected to the same preprocessing as for the Complete Works of Shakespeare. 
\add{To deal with the large vocabulary size of the Chinese characters, the 
model was trained at the byte level \citep{sennrich16} for Hong Lou Meng, resulting in 
a text of 2,932,451 bytes.} To measure the rank-frequency distribution and vocabulary 
growth at the word level, the model had to learn not only the sequence of bytes but also 
the splits between them.

The observations made for the Complete Works of Shakespeare apply also to
\figref{PTB:zipfheaps} and \figref{Chinese:zipfheaps}.
We observe power laws for both the rank-frequency distributions and the
vocabulary growth. The stacked LSTM replicates the power-law behaviors
well, reproducing approximately the same shapes for smaller $n$-grams.
The intersection of the uni-gram and 2-gram rank-frequency
distributions is reproduced as well. As for the vocabulary growth, the
reproduced exponents were a little larger than the original values, as
seen for the case of Shakespeare.

\figref{PTB:zipfheaps} also highlights the high capacity of the
stacked LSTM in learning with long $n$-grams. The top right graph
demonstrates that the stacked LSTM could repeat the same expression of
8-grams and 16-grams obeying Zipf's law.  In the Complete Works of
Shakespeare, written by a single author, long repeated $n$-grams
hardly occur, but the WSJ dataset contains many of these. For the
%unprocessed 
WSJ data, the rank-frequency distributions of 8- and
16-grams do not obviously follow power laws, mainly because of
repetition of the same expressions.  With such a corpus, the stacked
LSTM can also reproduce the power-law behavior of the rank-frequency
distribution of long $n$-grams.

These results indicate that a neural language model can learn the statistical laws behind 
natural language, and that the stacked LSTM is especially capable of reproducing both 
patterns of $n$-grams and the properties of vocabulary growth.

We also tested language models with different architectures. S1Fig shows results with 
different neural architectures for pseudo-texts generated for the Complete Works of Shakespeare: 
a convolutional neural net (CNN), simple RNN, single-layer LSTM, and stacked LSTM. 
\add{The stacked LSTM model was explained in \secref{sec:stacked}, while the details of the 
other models are given in the caption of S1Fig.} The two bottom right graphs for the 
stacked LSTM are identical to the two righthand graphs in 
\figref{shakespeare:zipfheaps}. Overall, all the models using an RNN reproduce power 
law behavior, but a closer look reveals greater capacity with the stacked LSTM. With 
the CNN (upper left), on the other hand, the shape of the rank-frequency distribution is 
quite different, and the exponent of the vocabulary growth is too large. The simple RNN 
(upper right) shows weaker capacity in reproducing longer $n$-grams, and the exponent 
is still too large. Finally, the single-layer LSTM (bottom left) is less capable of learning 
the longest $n$-gram of 5-grams as compared with the stacked LSTM (bottom right). 

\begin{figure}[t]
\begin{center}
\includegraphics[width=\textwidth]{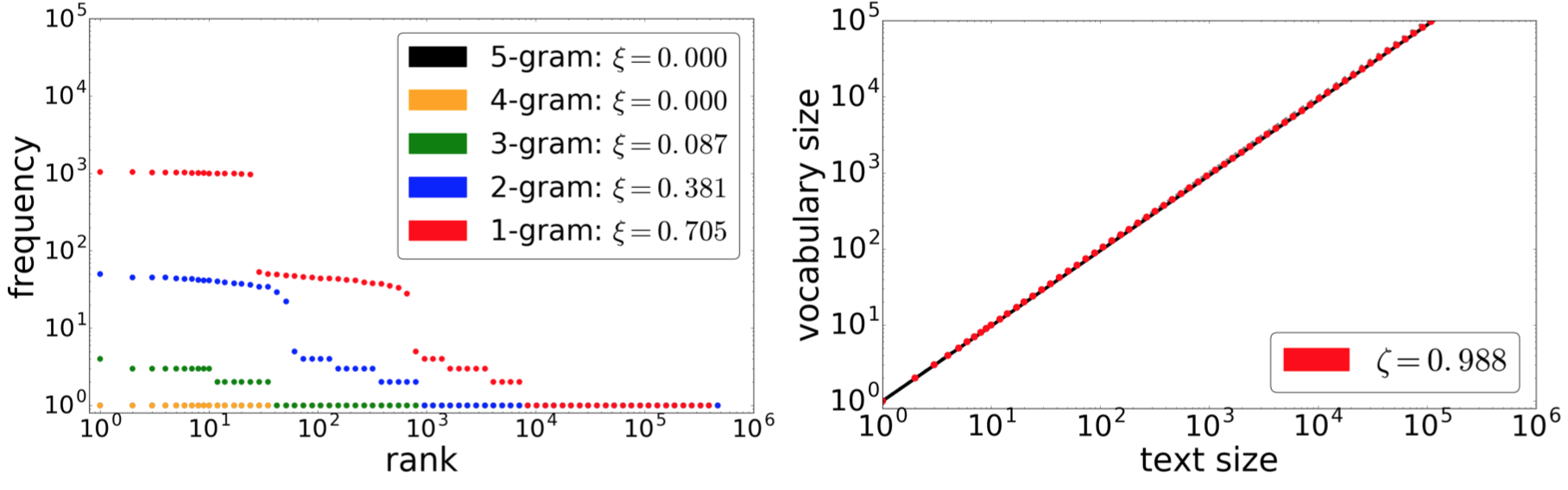}
\caption{Rank-frequency distribution (left) and vocabulary growth (right) for a text
\add{of 20 million characters} generated by the stacked LSTM without learning.}
\label{nolearning}
\end{center}
\end{figure}

\section{The Emergence of Zipf's Law and Heaps' Law}
The stacked LSTM acquires the behaviors of Zipf's law and Heaps' law
as learning progresses. It starts learning obviously at the level of a
monkey typing.  \figref{nolearning} shows the rank-frequency
distribution and vocabulary growth of a texts generated by the stacked
LSTM without training. Each case from uni-grams to 3-grams roughly
forms a power-law kind of step function. The vocabulary growth follows
a power law with exponent $\zeta \approx 1$ because monkey typing
consistently generates ``new words.''

\begin{figure}[t]
\begin{center}+1
\includegraphics[width=\textwidth]{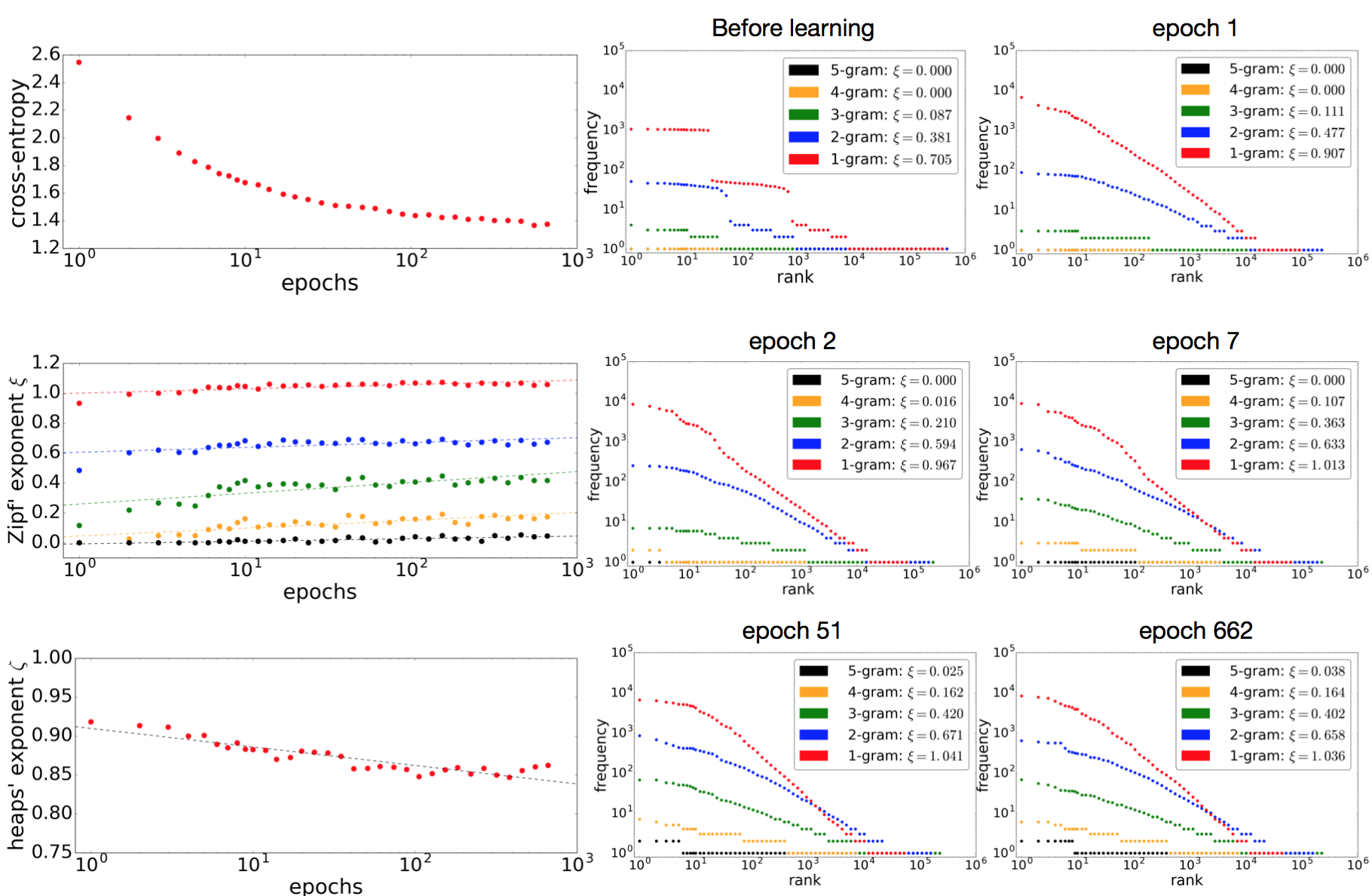}
\caption{\add{Training} cross entropy as a function of the number of epochs (upper 
left)\add{; Zipf's exponent $\xi$ (middle left)} and Heaps' exponent $\zeta$ (lower 
left) of pseudo-texts generated at different epochs; and the rank-frequency distributions 
of the pseudo-texts at various epochs (right) for the Complete Works of Shakespeare. The 
left-hand graphs are in logarithmic scale for the x-axes and linear scale for the y-axes. 
The fitting lines for Zipf's exponent $\xi$ and Heaps' exponent $\zeta$ were estimated 
from a linear regression with a semi-log scale.}
\label{dl_development}
\end{center}
\end{figure}

As shown by \figref{nolearning}, monkey-typed texts can theoretically 
produce power-law-like behaviors in the rank-frequency distribution and vocabulary 
growth. \citep{miller57} demonstrates how monkey typing generates a power-law 
rank-frequency distribution. Following the explanation in \citep{mitzenmacher03}, we 
briefly summarize the rationale as follows. Consider a monkey that randomly types any 
of $n$ characters and the space bar. Since a space separates words, let its probability be 
$q$, and then each of the other characters is hit uniformly with a probability of 
$(1-q)/n$. Given that the number of words of length $c$ is $n^c$, and that longer words 
are less likely to occur, then the rank frequency of a word of length $c$ is between 
$S(c)+1$ and $S(c+1)$, where $S(c) = \sum_{i=1}^c n^i$. Since $S(c) = \frac{n^c - 
1}{n-1}$, the rank $r_c$ of a word of length $c$ grows exponentially with respect to 
$c$; i.e., $r_c \approx n^c$. Given that the probability of occurrence of a word of 
length $c$ is $q (\frac{1-q}{n})^c$, by replacing $c$ with the rank, we obtain the 
rank-probability distribution as
\begin{eqnarray}
  P (r_c) = q (\frac{1-q}{n})^{\log r_c} = q (r_c)^{\log  (1-q)-1},
\end{eqnarray}
where the $\log$ is taken with base $n$. This result shows that the probability 
distribution follows a power law with respect to the rank. The LSTM models therefore 
start learning by innately possessing a power-law feature for the rank-frequency 
distribution and vocabulary growth. The learning process thus smooths the step-like 
function into a more continuous distribution; moreover, it decreases the exponent for 
vocabulary growth. \citep{bcw} reports empirically that when the probabilities of each 
character are different, the rank-frequency distribution becomes smoother. While 
learning progresses, the exponent $\zeta$ is lowered by learning patterns within texts.

\figref{dl_development} illustrates the training progress of the language model for the 
Complete Works of Shakespeare. The upper-left graph shows the cross entropy of the model at 
different \add{training} epochs. The training successfully decreases the cross entropy 
and reaches a convergent state.

\add{The middle and lower left graphs in \figref{dl_development} show the Zipf's 
exponent $\xi$ and Heaps' exponent $\zeta$ of the pseudo-texts generated at different 
epochs. At the very beginning of training, the Zipf's exponents $\xi$ tend to be smaller 
than that of the original dataset. They generally increase and become equivalent to the 
values of the original datasets for short $n$-grams or remain at smaller values for long 
$n$-grams.} As the model minimizes the cross entropy, the Heaps' exponent 
$\zeta$ generally decreases, with some fluctuation, by learning words. It roughly stops 
decreasing, however, at around $10^2$ to$10^3$ epochs. The fact that the exponents of 
Heaps' law cannot reach the value of the original text indicates some limitation in 
learning.

The right-hand side of \figref{dl_development} shows the rank-frequency distributions 
of the pseudo-texts generated at different epochs. The stacked LSTM model reproduces 
the power-law behavior well for uni-grams and 2-grams, and partially for 3-grams, with 
just a single epoch (upper right). Such behavior for 4-grams appears in epoch 2 (middle 
left), and the intersection of the uni-gram and 3-gram power laws appears in epoch 7 
(middle right). Power-law behavior for 5-grams emerges in epoch 51 (bottom left), and 
no further qualitative change is observed afterwards (bottom right).

As training progresses, the stacked LSTM first learns short patterns (uni-grams and 
2-grams) and then gradually acquires longer patterns (3- to 5-grams). It also learns 
vocabulary as training progresses, which lowers the exponent of Heaps' law. There are 
no tipping points at which the neural nets drastically change their behavior, and the two 
power laws are both acquired at a fairly early stage of learning.

\section{Neural language models are limited in reproducing 
long-range correlation}
\label{sec:longrange}
Natural language has structural features other than $n$-grams that underlie the 
arrangement of words. A representative of such features is grammar, which has been 
described in various ways in the linguistics domain. The structure underlying the 
arrangement of words has been reported to be scale-free, globally ranging across 
sentences and at the whole-text level. One quantification methodology for such global 
structure is long-range correlation.

Long-range correlation describes a property by which two subsequences within a 
sequence remain similar even with a long distance between them. Typically, such 
sequences have a power-law relationship between the distance and the similarity. This 
statistical property is observed for various sequences in complex systems. Various 
studies 
\citep{Ebeling1994,Ebeling1995,Montemurro2002,Kosmidis2006,Altmann2009,Altmann2012, Montemurro2014, plos16} report that natural language has long-range 
correlation as well.

\subsection{The power decay of mutual information is unlikely to hold for natural 
language text}
Measurement of long-range correlation is not a simple problem, as we will see, and 
various methods have been proposed. \citep{Lin_2016} proposes applying mutual 
information to measure long-range dependence between symbols. The mutual 
information at a distance $s$ is defined as
%\begin{equation}
%  I_s (X,Y) = \sum_{a,b} P (a,b) \log \frac{P (a,b)}{P (a)P (b)},
%\end{equation}
\begin{equation}
 \add{I_s (X,Y) = \sum_{X=a,Y=b} P (X,Y) \log \frac{P (X,Y)}{P (X)P (Y)}}
\end{equation}

where $X$ and $Y$ are random variables of elements in each of two subsequences at 
distance $s$.

\citep{Lin_2016} proves mathematically how a sequence generated with their 
\add{simple recursive grammar model} results in power decay of the mutual 
information. They also provide empirical evidence that a Wikipedia source from the 
enwik8 dataset exhibits power decay of the mutual information, and that a pseudo-text 
generated from Wikipedia also exhibits power decay when measured at the character 
level. The left graph in Fig. 6 shows our reproduction of the mutual information for the 
Wikipedia source used in \citep{Lin_2016}. For each of the graphs in the figure, the 
horizontal axis indicates the distance $s$, and the vertical axis indicates the mutual 
information. The red and blue points represent the results for the real texts and the 
pseudo-texts, respectively. By using the data in \citep{Lin_2016}, we could reproduce 
their results: for the Wikipedia source, the mutual information exhibits power decay, and 
this statistical property was also learned by the stacked LSTM.

\begin{figure}[t]
\begin{center}
\includegraphics[width=\textwidth]{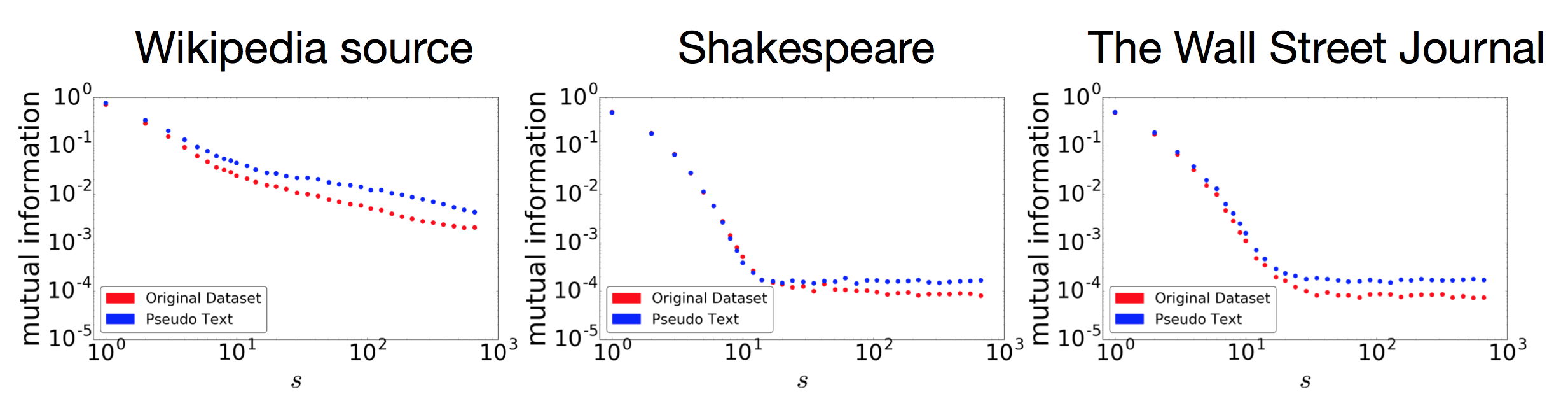}
\end{center}
\caption{Mutual information for a Wikipedia source (left), the Complete Works of Shakespeare 
(middle), and The Wall Street Journal (right). The red and blue points represent the 
mutual information of the datasets and the generated pseudo-texts, respectively. 
Following \citep{Lin_2016}, the Wikipedia {\em source} was preprocessed to exclude 
rare symbols.}
\label{long-range:lin}
\end{figure}

We doubt, however, that the power decay of the mutual information is being properly 
observed for a natural language text when measured with the method proposed in 
\citep{Lin_2016}. The middle and right graphs in Fig. 6 show the results for the 
Complete Works of Shakespeare and The Wall Street Journal, which are more standard 
natural language datasets. The mutual information exhibits an exponential decay 
showing short-range correlation similar to what they reported as the behavior of Markov 
models, and it almost reaches a plateau just after a 10-character distance for both 
datasets. This plateau represents a state in which the probabilistic distributions of 
$a$ and $b$ pairs become almost the same because of the low frequency problem. 
Following \add{the statistical properties of the datasets}, \add{the} 
stacked LSTM replicates this exponential decay well. 
\citep{Lin_2016} also examines a natural language corpus, corpatext. Unfortunately, 
corpatext is poorly organized, as it contains a huge number of repeats of long $n$-grams, 
chains of numbers, many meta-characters, and successive spaces. Our measurement of 
mutual information with corpatext (which was preprocessed to delete meta-characters 
and successive spaces) gave an exponential decay up to 10 characters. This observation 
is similar to the results for the Complete Works of Shakespeare and WSJ. The mutual information 
slowly decayed after a length of 10, however, which could lead to misinterpretation of 
the power decay. \citep{Lin_2016} does not clearly mention any preprocessing or 
measurement of the mutual information of a pseudo-text for corpatext. We cannot reach 
a solid conclusion with such an unorganized corpus.

There are two reasons for this difference in results between the Wikipedia source and 
the Shakespeare/WSJ: the kind of data, and the quantification method. Regarding the 
kind of data, we must emphasize that \citep{Lin_2016} does not use any standard 
natural language text for verification. Instead, they use the wiki {\em source}, including 
all Wikipedia annotations. Therefore, Wikipedia is strongly grammatical,
that they consider for their mathematical proof.

As for the problem of the quantification method, as seen from the plateau appearing in 
the results for the Shakespeare and WSJ datasets, the mutual information in its basic 
form is highly susceptible to the low frequency problem. Therefore, \citep{Lin_2016} 
verifies data with a small alphabet size (including DNA sequences). When the alphabet 
is increased to the size of the Chinese character set, the mutual information reaches the 
plateau almost immediately; when using words, the plateau is reached in only two or 
three points.

Still, \citep{Lin_2016} provides a crucial understanding of neural nets: the stacked 
LSTM (or other LSTM-like models) can replicate the power decay of the mutual 
information, if it exists in the original data. Whether such strong long-range correlation 
exists, however, depends on the data type. Given all other reports, as will be mentioned 
shortly, long-range correlation {\em does} exist in natural language texts. The problem 
of how to quantify it is non-trivial, however, and the mutual information, as proposed, is 
not always a good measurement for natural language.

\begin{figure}[t]
\begin{center}
\includegraphics[width=\textwidth]{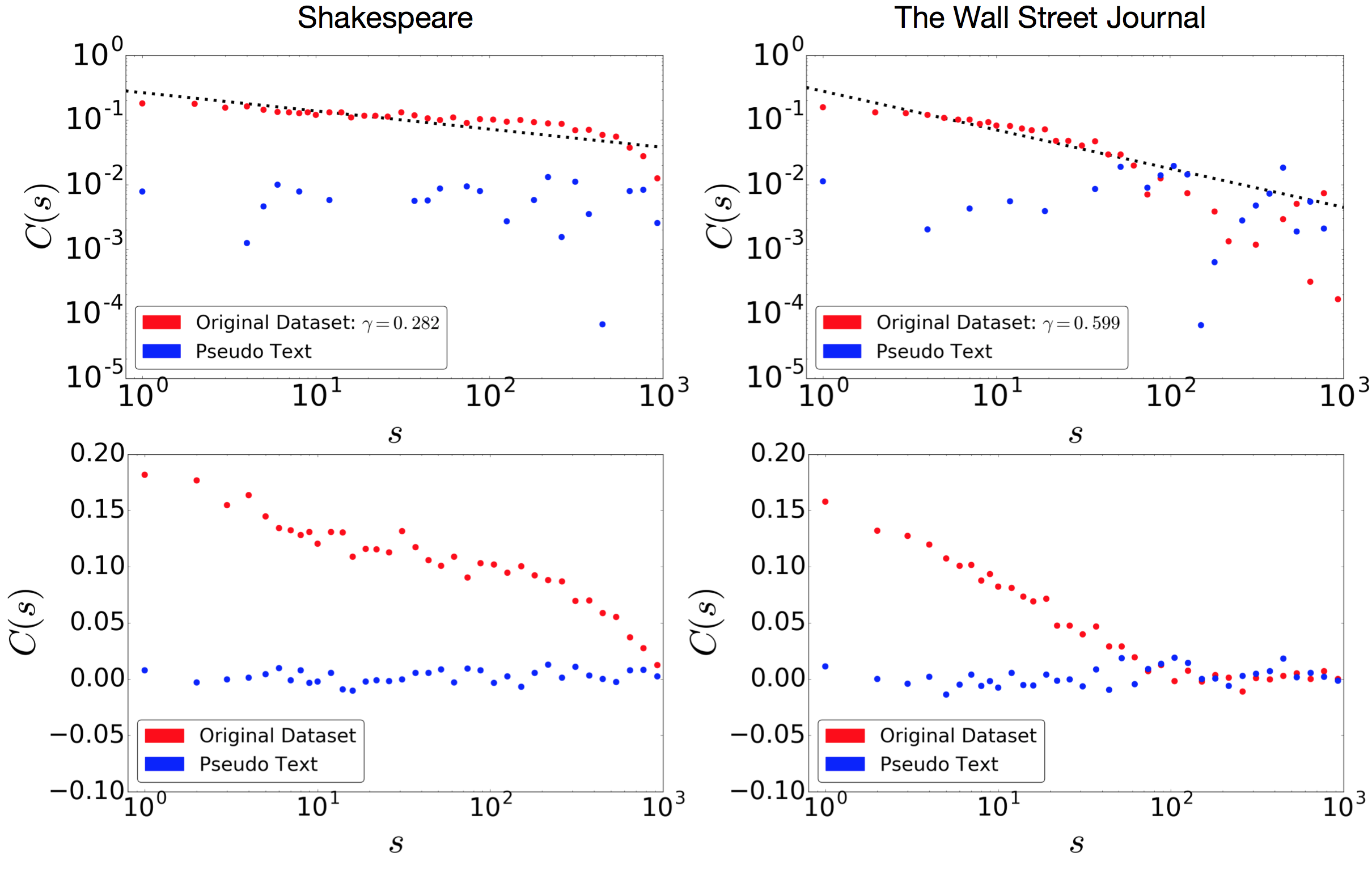}
\caption{Long-range correlation measured with the autocorrelation function by the 
method of \citep{plos16} for the original and generated texts of the Complete Works of 
Shakespeare (left) and The Wall Street Journal (right). The upper and lower graphs are 
in log-log and semi-log scale, respectively. The fitting line for The Wall Street Journal 
was estimated from the data points where $s < 100$.}
\label{long-range:interval}
\end{center}
\end{figure}

\subsection{Neural language models cannot reproduce the power decay of the 
autocorrelation function in natural language}
\label{sec:acf}
Quantification of long-range correlation has been studied in the statistical physics 
domain and has been effective in analyzing extreme events in natural phenomena and 
financial markets \citep{Corral1, C4, Bunde2005, Kantz, 
Blender2015,Turcotte,Yamasaki,Bogachev}. The long-range correlation in this 
application is also a scale-free property taking a power-law form. Long-range 
correlation is explained here as a ``clustering phenomenon'' of rare events. This could 
have some relation to an underlying grammar-like structure in a sequence, but the 
measure might quantify a phenomenon different from that captured by the mutual 
information.

The application of long-range correlation to natural language is controversial, because 
all proposed methods are for numerical data, whereas natural language has a different 
nature. Various reports show how natural language is indeed long-range correlated 
\citep{Ebeling1994,Ebeling1995,Montemurro2002,Altmann2009,Altmann2012,Montemurro2014}. We apply the most recent method \citep{plos16} to investigate 
whether the pseudo-text generated by a stacked LSTM retains long-range correlation. 

This method is based on the autocorrelation function applied to a sequence 
$R=r_1,r_2,\ldots,r_N$ with mean $\mu$ and standard deviation $\sigma$:
\begin{equation}
 C(s)= \frac{1}{ (N-s)\sigma^2}\sum_{i=1}^{N-s} (r_i-\mu) (r_{i+s}-\mu),
\end{equation}
with $C(0)=1.0$ by definition. Note that this function $C(s)$ measures the similarity 
between two subsequences that are distance $s$ apart. If this autocorrelation function 
exhibits a power decay as follows, \begin{equation} C(s)=C (1)s^{-\gamma}, s>0, 
\label{eq:long-range-correlation} \end{equation} then the sequence $R$ is long-range 
correlated. The functional range of $C(s)$ is between $-1.0$ and $1.0$. If there is no 
correlation, $C(s)$ is almost zero; if the sequences are well correlated positively, 
$C(s)$ takes a positive value. The method used here is based on the intervals 
between rare words, and we use the rarest 1/16th of words among all words appearing 
in a text, following \citep{plos16}, in defining the interval sequence $R$.

\figref{long-range:interval} shows the autocorrelation functions of
the Complete Works of Shakespeare (left) and The Wall Street Journal (right)
at the word level. The upper graphs show log-log plots, whereas the
lower graphs show semi-log plots for the same sets of results.  The
red and blue points represent the original dataset and the
pseudo-text, respectively. For the original Complete Works of Shakespeare,
the results exhibit a clear, slow power decay up to $s=10^{3}$.  This
behavior is similar to that of other literary texts reported in
\citep{plos16}. In contrast, the autocorrelation function of the
pseudo-text takes values around $0$ for any $s$. The Wall Street
Journal results show similar behavior. The power decay is faster than
that of the Complete Works of Shakespeare and other single-author literary
texts, but the autocorrelation function still takes positive values
and exhibits power decay up to about $s=10^2$.

In summary, this analysis provides qualitative evidence regarding a shortcoming of the 
stacked LSTM: it has a limitation with respect to reproducing long-range correlation, as 
quantified using a method proposed in the statistical physics domain.

\add{To further clarify the stacked LSTM's performance and limitation, we conducted 
three experiments from different perspectives. First, S2Fig shows the encoding rate 
decay of the original text (WSJ) with different types of shuffling and the pseudo-text. 
The encoding rate is a measure of a text's predictability, and we provide more 
explanation in the caption of S2Fig. The results show that the pseudo-text generated by 
the stacked LSTM model is more predictable than the word-shuffled WSJ, and less 
predictable than both the original and document-shuffled WSJ.}

\add{Second, S3Fig shows the mutual information and autocorrelation function of the 
original text (Shakespeare) and pseudo-texts generated by different neural architectures. 
The mutual information decays rapidly up to around $s=10$, and the models except for 
the CNN model reproduce the behavior of the mutual information of the original dataset 
well. On the other hand, the power decay exhibited by the original dataset was never 
reproduced by any model that we tested.}

\add{Third, S4Fig shows the autocorrelation function for the French novel, Les 
Mis\'erables, and the text translated into English by a neural machine translation system. 
We obtained the translated text from the Google Cloud Translation API({\tt 
https://cloud.google.com/translate/}). The translated text maintains the power decay of 
the autocorrelation function observed for the original text. This can be explained from 
the properties of machine translation. Machine translation is not expected 
to radically change the order of corresponding words between sentences. Therefore, as 
long as the translation system has the capacity to output rare words in the original text, 
the autocorrelation should be preserved.}.

Because long-range correlation is a global, scale-free property of a text, one reason for 
the limitation of the stacked LSTM could lie in the context length of $k=128$ at the 
character level. Considering the availability of computational resources, however, this 
setting was a maximum limit, as the number of layers to be computed substantially 
increases with the context length. Moreover, it has been empirically reported that an 
LSTM \add{architecture} cannot retain past information beyond a length of 
$100$ \citep{Hochreiter_1997}.

One possible future approach is to test new neural models with enhanced long-memory 
features, such as a CNN application \citep{Zen_2016} \citep{Kalchbrenner_2016} or the 
hierarchical structure of an RNN \citep{Hihi_1995}\citep{Mehri_2016}. Overall, the 
behavior of pseudo-texts with respect to the statistical laws of natural language partly 
reveals both the effectiveness and limitations of neural networks, which tend to remain 
black boxes because of their complexity. Analysis using statistical laws could provide a 
direction towards improving the architectures of neural networks.

\section{Conclusion}
To understand the effectiveness and limitations of deep learning for natural language 
processing, we empirically analyzed the capacity of neural language models in terms of 
the statistical laws of natural language. This paper considered three statistical laws of 
natural language: Zipf's law, the power law underlying the rank-frequency distribution; 
Heaps' law, the power-law increase in vocabulary size with respect to text size; and 
long-range correlation, which captures the self-similarity underlying natural language 
sequences.

The analysis revealed that neural language models satisfy Zipf's law, not only for 
uni\add{-}grams, but also for longer $n$-grams. To the best of our knowledge, this is 
the first language model that can reproduce a statistical law at such a level. The 
language models also satisfy Heaps' law: they generate text with power-law vocabulary 
growth. The exponent remained higher than for the original texts, however, which 
showed both the limitation of detecting words and the self-organization of linguistic 
sequences.

Finally, a stacked LSTM showed a limitation with respect to capturing the long-range 
correlation of natural language. Investigation of a previous 
work\add{\citep{Lin_2016}} revealed that if the original learning text has a strong 
grammatical structure, then a stacked LSTM has the potential to reproduce it. A 
standard natural language text, however, does not have such a feature. The long-range 
correlation quantified with another methodology for the original texts was not 
reproduced by the stacked LSTM.

Our analysis suggests \add{a direction for improving language models, which has 
always been the central problem in handling natural language on machines. The current 
neural language models are unable to handle the global structures underlying texts. 
Because the Zipf's law behavior with long $n$-grams was reproduced well by the 
stacked LSTM, this neural language model has high capacity for recognizing and 
reproducing local patterns or phrases in the original text. The model could not, however, 
reproduce the long-range correlation measured by the autocorrelation function for 
intervals between rare words. This long-range correlation cannot be reduced to local 
patterns but rather is a representation of global structure in natural language. This 
irreproducibility demonstrates the limitation of neural language models and is a 
challenge for language modeling with neural networks.}

Our future work thus includes exploring conditions to reproduce the long-range 
correlation in text with language models, including both stochastic and neural language 
models. 

\bibliographystyle{natbib}
\bibliography{plos}

\begin{thebibliography}{}

\bibitem[Altmann {\em et~al.}(2009)]{Altmann2009}
Altmann, E.G. , Pierrehumbert, J.B. , and Motter, E.A.  (2009).
\newblock Beyond word frequency: Bursts, lulls, and scaling in the temporal
  distributions of words.
\newblock {\em PLOS one}.

\bibitem[Altmann {\em et~al.}(2012)]{Altmann2012}
Altmann, E.G. , Cristadoro, G. , and Esposti, M.D.  (2012).
\newblock On the origin of long-range correlations in texts.
\newblock In {\em Proceedings of the National Academy of Sciences}, volume 109,
  pages 11582--11587.

\bibitem[Baeza–Yates and Navarro(2000)]{BaezaYates_2000}
Baeza–Yates, Ricardo  and Navarro, Gonzalo  (2000).
\newblock Block addressing indices for approximate text retrieval.
\newblock {\em Journal of the American Society for Information Science}, {\bf
  51}(1), 69–82.

\bibitem[Bell {\em et~al.}(1990)]{bcw}
Bell, T.C. , Cleary, J.G. , and Witten, H.  (1990).
\newblock {\em Text Compression}.
\newblock Prentice-Hall.

\bibitem[Bengio {\em et~al.}(2003)]{bengio03}
Bengio, Y. , Ducharme, R. , Vincent, P. , and Jauvin, C.  (2003).
\newblock A neural probabilistic language model.
\newblock {\em The Journal of Machine Learning Research}, {\bf 3}, 1137--1155.

\bibitem[Bianchini and Scarselli(2014)]{bianchini2014complexity}
Bianchini, Monica  and Scarselli, Franco  (2014).
\newblock On the complexity of neural network classifiers: A comparison between
  shallow and deep architectures.
\newblock {\em IEEE transactions on neural networks and learning systems}, {\bf
  25}(8), 1553--1565.

\bibitem[Blender {\em et~al.}(2015)]{Blender2015}
Blender, R. , Raible, C. , and Lunkeit, F.  (2015).
\newblock Non-exponential return time distributions for vorticity extremes
  explained by fractional poisson processes.
\newblock {\em Quarterly Journal of the Royal Meteorology Society}, {\bf 141},
  249--257.

\bibitem[Bogachev {\em et~al.}(2007)]{Bogachev}
Bogachev, M.I. , Eichner, J.F. , and Bunde, A.  (2007).
\newblock Effect of nonlinear correlations on the statistics of return
  intervals in multifractal data sets.
\newblock {\em Physical Review Letters}, {\bf 99}(240601).

\bibitem[Bunde {\em et~al.}(2005)]{Bunde2005}
Bunde, A. , Eichner, J. , Havlin, S. , and Kantelhardt, J.W.  (2005).
\newblock Long-term memory: A natural mechanism for the clustering of extreme
  events and anomalous residual times in climate records.
\newblock {\em Physical Review Letters}, {\bf 94}(048701).

\bibitem[Chelba {\em et~al.}(2017)]{Chelba_2017}
Chelba, Cipran , Norouzi, Mohammad , and Bengio, Samy  (2017).
\newblock N-gram language modeling using recurrent neural network estimation.
\newblock {\em arXiv preprint}, {\bf abs/1703.10724}.

\bibitem[Cohen and Shashua(2016)]{ICML:Cohen+Shashua:2016}
Cohen, Nadav  and Shashua, Amnon  (2016).
\newblock Convolutional rectifier networks as generalized tensor
  decompositions.
\newblock In {\em Proceedings of the 33th International Conference on Machine
  Learning}, pages 955--963.

\bibitem[Cohen {\em et~al.}(2016)]{cohen2016expressive}
Cohen, Nadav , Sharir, Or , and Shashua, Amnon  (2016).
\newblock On the expressive power of deep learning: A tensor analysis.
\newblock In {\em Proceedings of the 29th Annual Conference on Learning
  Theory}, pages 698--728.

\bibitem[Corral(1994)]{Corral1}
Corral, A.  (1994).
\newblock Long-term clustering, scaling, and universality in the temporal
  occurrences of earthquakes.
\newblock {\em Physical Review Letters}, {\bf 92}(108501).

\bibitem[Corral(2005)]{C4}
Corral, A.  (2005).
\newblock Renomalization-group transformations and correlations of seismicity.
\newblock {\em Physical Review Letters}, {\bf 95}(028501).

\bibitem[Ebeling and Neiman(1995)]{Ebeling1995}
Ebeling, W.  and Neiman, A.  (1995).
\newblock Long-range correlations between letters and sentences in texts.
\newblock {\em Physica A}, {\bf 215}, 233--241.

\bibitem[Ebeling and P\"oschel(1994)]{Ebeling1994}
Ebeling, W.  and P\"oschel, T.  (1994).
\newblock Entropy and long-range correlations in literary english.
\newblock {\em Europhysics Letters}, {\bf 26}, 241--246.

\bibitem[Guiraud(1954)]{Guiraud1954}
Guiraud, H.  (1954).
\newblock {\em Les Charact\`eres Statistique du Vocabulaire}.
\newblock Universitaires de France Press.

\bibitem[Gulcehre {\em et~al.}(2016)]{acl14}
Gulcehre, C. , Ahn, S. , Nallapati, R. , Hou, B. , and Bengio, Y.  (2016).
\newblock Pointing the unknown words.
\newblock pages 140--149.

\bibitem[Heaps(1978)]{heaps}
Heaps, H.~S.  (1978).
\newblock {\em Information Retrieval: Computational and Theoretical Aspects,
  Academic Press}.

\bibitem[Herdan(1964)]{Herdan1964}
Herdan, G.  (1964).
\newblock {\em Quantitative Linguistics}.
\newblock Butterworths.

\bibitem[Hihi and Bengio(1995)]{Hihi_1995}
Hihi, Salah  and Bengio, Yoshua  (1995).
\newblock Hierarchical recurrent neural networks for long-term dependencies.
\newblock In {\em Advances in Neural Information Processing Systems 8}, pages
  493--–499.

\bibitem[Hilberg(1990)]{Hilberg}
Hilberg, W.  (1990).
\newblock {Der bekannte Grenzwert der redundanzfreien Information in Texten ---
  eine Fehlinterpretation der Shannonschen Experimente?}
\newblock {\em Frequenz}, {\bf 44}, 243--248.

\bibitem[Hochreiter and Schmidhuber(1997)]{Hochreiter_1997}
Hochreiter, S  and Schmidhuber, J  (1997).
\newblock Long short-term memory.
\newblock {\em Neural computation}, {\bf 9}(8), 1735–1780.

\bibitem[Kalchbrenner {\em et~al.}(2016)]{Kalchbrenner_2016}
Kalchbrenner, Nam , Espeholt, Lasse , Simonyan, Karen , van~den Oord, Aaron ,
  Graves, Alex , and Kavukcuoglu, Koray  (2016).
\newblock Neural machine translation in linear time.
\newblock {\em arXiv preprint}, {\bf abs/1610.10099}.

\bibitem[Kingma and Ba(2014)]{Kingma_2014}
Kingma, D  and Ba, J  (2014).
\newblock Adam: A method for stochastic optimization.
\newblock {\em arXiv preprint}, {\bf abs/1412.6980}.

\bibitem[Kosmidis {\em et~al.}(2012)]{Kosmidis2006}
Kosmidis, K. , Kalampokis, A. , and Argyrakis, K.  (2012).
\newblock Language time series analysis.
\newblock {\em Physica A}, {\bf 370}, 808--816.

\bibitem[Krause {\em et~al.}(2016)]{Krause}
Krause, B. , Lu, L. , Murray, I. , and Renals, S.  (2016).
\newblock Multiplicative lstm for sequence modeling.
\newblock {\em arXiv preprint}, {\bf abs/1609.07959}.

\bibitem[Lin and Tegmark(2016a)]{Lin_2016}
Lin, H  and Tegmark, M  (2016a).
\newblock Critical behavior from deep dynamics: A hidden dimension in natural
  language.
\newblock {\em arXiv preprint}, {\bf abs/1606.06737}.

\bibitem[Lin and Tegmark(2016b)]{Lin_Tegmark_2016}
Lin, Henry  and Tegmark, Max  (2016b).
\newblock Why does deep and cheap learning work so well?
\newblock {\em arXiv preprint}, {\bf abs/1608.08225}.

\bibitem[Lu {\em et~al.}(2010)]{lu2010}
Lu, Linyuan , Zhang, Zi-Ke , and Zhou, Tao  (2010).
\newblock Zipf's law leads to heaps' law: Analyzing their relation in
  finite-size systems.
\newblock {\em arXiv preprint}.

\bibitem[Luong and Manning(2016)]{Luong_2016}
Luong, Minh-Thang  and Manning, Christopher  (2016).
\newblock Achieving open vocabulary neural machine translation with hybrid
  word-character models.
\newblock In {\em Proceedings of the 54th Annual Meeting of the Association for
  Computational Linguistics}, page 1054–1063.

\bibitem[Mehri {\em et~al.}(2016)]{Mehri_2016}
Mehri, Soroush , Kumar, Kundan , Gulrajani, Ishaan , Kumar, Rithesh , Jain,
  Shubham , Sotelo, Jose , Courville, Aaron , and Bengio, Yoshua  (2016).
\newblock Samplernn: An unconditional end-to-end neural audio generation model.
\newblock {\em arXiv preprint}, {\bf abs/1612.07837}.

\bibitem[Miller(1957)]{miller57}
Miller, G.A.  (1957).
\newblock Some effects of intermittent silence.
\newblock {\em American Journal of Psychology}, {\bf 70}, 311--314.

\bibitem[Mitzenmacher(2003)]{mitzenmacher03}
Mitzenmacher, M.  (2003).
\newblock A brief history of generative models for power law and lognormal
  distributions.
\newblock {\em Internet Mathematics}, {\bf 1}(2), 226--251.

\bibitem[Montemurro(2014)]{Montemurro2014}
Montemurro, M.A.  (2014).
\newblock Quantifying the information in the long-range order of words:
  Semantic structures and universal linguistic constraints.
\newblock {\em Cortex}, {\bf 55}, 5--16.

\bibitem[Montemurro and Pury(2002)]{Montemurro2002}
Montemurro, M.  and Pury, P.A.  (2002).
\newblock Long-range fractal correlations in literary corpora.
\newblock {\em Fractals}, {\bf 10}, 451--461.

\bibitem[Montufar {\em et~al.}(2014)]{NIPS2014_5422}
Montufar, Guido~F. , Pascanu, Razvan , Cho, Kyunghyun , and Bengio, Yoshua
  (2014).
\newblock On the number of linear regions of deep neural networks.
\newblock In {\em Advances in Neural Information Processing Systems 27}, pages
  2924--2932.

\bibitem[Pitman(2006)]{pitman}
Pitman, J.  (2006).
\newblock {\em Combinatorial Stochastic Processes}.
\newblock Springer.

\bibitem[Poole {\em et~al.}(2016)]{NIPS:Poole+etal:2016}
Poole, Ben , Lahiri, Subhaneil , Raghu, Maithreyi , Sohl-Dickstein, Jascha ,
  and Ganguli, Surya  (2016).
\newblock Exponential expressivity in deep neural networks through transient
  chaos.
\newblock In {\em Advances in Neural Information Processing Systems 29}, pages
  3360--3368.

\bibitem[Rush {\em et~al.}(2015)]{Rush:EMNLP:2015}
Rush, Alexander~M. , Chopra, Sumit , and Weston, Jason  (2015).
\newblock A neural attention model for abstractive sentence summarization.
\newblock In {\em Proceedings of the 2015 Conference on Empirical Methods in
  Natural Language Processing}, pages 379--389.

\bibitem[Santhanam and Kantz(2005)]{Kantz}
Santhanam, M.  and Kantz, H.  (2005).
\newblock Long-range correlations and rare events in boundary layer wind
  fields.
\newblock {\em Physica A}, {\bf 345}, 713--721.

\bibitem[{Schwab} and {Mehta}(2016)]{Schwab_2016}
{Schwab}, D.~J.  and {Mehta}, P.  (2016).
\newblock {Comment on ``Why does deep and cheap learning work so well?''}.
\newblock {\em arXiv preprint}, {\bf 1abs/1609.03541}.

\bibitem[Sennrich {\em et~al.}(2016)]{sennrich16}
Sennrich, R. , Haddow, B. , and Birch, A.  (2016).
\newblock Neural machine translation of rare words with subword units.
\newblock In {\em Proceedings of the 54th Annual Meeting of the Association for
  Computational Linguistics}, pages 1715--1725.

\bibitem[Serban {\em et~al.}(2015)]{Serban:AAAI:2015}
Serban, Iulian , Sordoni, Alessandro , Bengio, Yoshua , Courville, Aaron , and
  Pineau, Joelle  (2015).
\newblock Building end-to-end dialogue systems using generative hierarchical
  neural network models.
\newblock In {\em Proceedings of the 30th AAAI Conference on Artificial
  Intelligence}.

\bibitem[Simon(1955)]{simon55}
Simon, H.A.  (1955).
\newblock On a class of skew distribution functions.
\newblock {\em Biometrika}, {\bf 42}(3/4), 425--440.

\bibitem[Sundermeyer {\em et~al.}(2012)]{Sundermeyer_2012}
Sundermeyer, Martin , Schlüter, Ralf , and Herman, Ney  (2012).
\newblock Lstm neural networks for language modeling.
\newblock In {\em 13th Annual Conference of the International Speech
  Communication Association}, pages 194--197.

\bibitem[Takahira {\em et~al.}(2016)]{entropy16}
Takahira, Ryosuke , Tanaka-Ishii, Kumiko , and Lukasz, Debowski  (2016).
\newblock Large scale verification of entropy of natural langauge.
\newblock {\em Entropy}.
\newblock Online Journal. HTML Version: {\tt
  http://www.mdpi.com/1099-4300/18/10/364/html}, PDF Version: {\tt
  http://www.mdpi.com/1099-4300/18/10/364/pdf}.

\bibitem[Tan {\em et~al.}(2015)]{Tan:ICLR:2015}
Tan, Ming , dos Santos, Cicero , Xiang, Bing , and Zhou, Bowen  (2015).
\newblock Lstm-based deep learning models for non-factoid answer selection.
\newblock {\em arXiv preprint}, {\bf abs/1511.04108}.

\bibitem[Tanaka-Ishii and Bunde(2016)]{plos16}
Tanaka-Ishii, K.  and Bunde, A.  (2016).
\newblock Long-range memory in literary texts: On the universal clustering of
  the rare words.
\newblock {\em PLOS One}, {\bf 11}(11), e0164658.

\bibitem[Turcotte(1997)]{Turcotte}
Turcotte, D.L.  (1997).
\newblock {\em Fractals and Chaos in Geology and Geophysics}.
\newblock Cambridge University Press.

\bibitem[van~den Oord {\em et~al.}(2016)]{Zen_2016}
van~den Oord, A , Dieleman, S , and Zen, H  (2016).
\newblock Wavenet: A generative model for raw audio.
\newblock {\em arXiv preprint}, {\bf abs/1609.03499}.

\bibitem[van Leijenhorst and van~der Weide(2005)]{Leijenhorst_2005}
van Leijenhorst, D.C.  and van~der Weide, Th.P.  (2005).
\newblock A formal derivation of heaps’ law.
\newblock {\em Information Sciences}, {\bf 170}(2-4), 263–272.

\bibitem[Wu {\em et~al.}(2016)]{Wu_2016}
Wu, Y , Schuster, M , Chen, Z , Le, Q , Norouzi, M , Macherey, W , Krikun, M ,
  Cao, Y , Gao, Q , Macherey, K , and et~al. (2016).
\newblock Google’s neural machine translation system: Bridging the gap
  between human and machine translation.
\newblock {\em arXiv preprint}, {\bf abs/1609.08144}.

\bibitem[Yamasaki {\em et~al.}(2007)]{Yamasaki}
Yamasaki, K. , Muchnik, L. , Havlin, S. , Bunde, A. , and Stanley, H.E.
  (2007).
\newblock Scaling and memory in volatility return intervals in financial
  markets.
\newblock {\em Proceedings of the National Acaddemy of Sciences}, {\bf 102},
  9424--9428.

\bibitem[Zipf(1965)]{zipf}
Zipf, G.K.  (1965).
\newblock {\em Human behavior and the principle of least effort: An
  introduction to human ecology}.
\newblock Hafner.

\end{thebibliography}
\newpage

We thank JST PRESTO for financial support. Moreover, we thank Ryosuke Takahira of 
the Tanaka-Ishii Group for his help in creating Fig. 6.
\newpage
\section*{Supporting Information}

\begin{figure}[h]
\begin{center}
\includegraphics[width=0.6\textwidth]{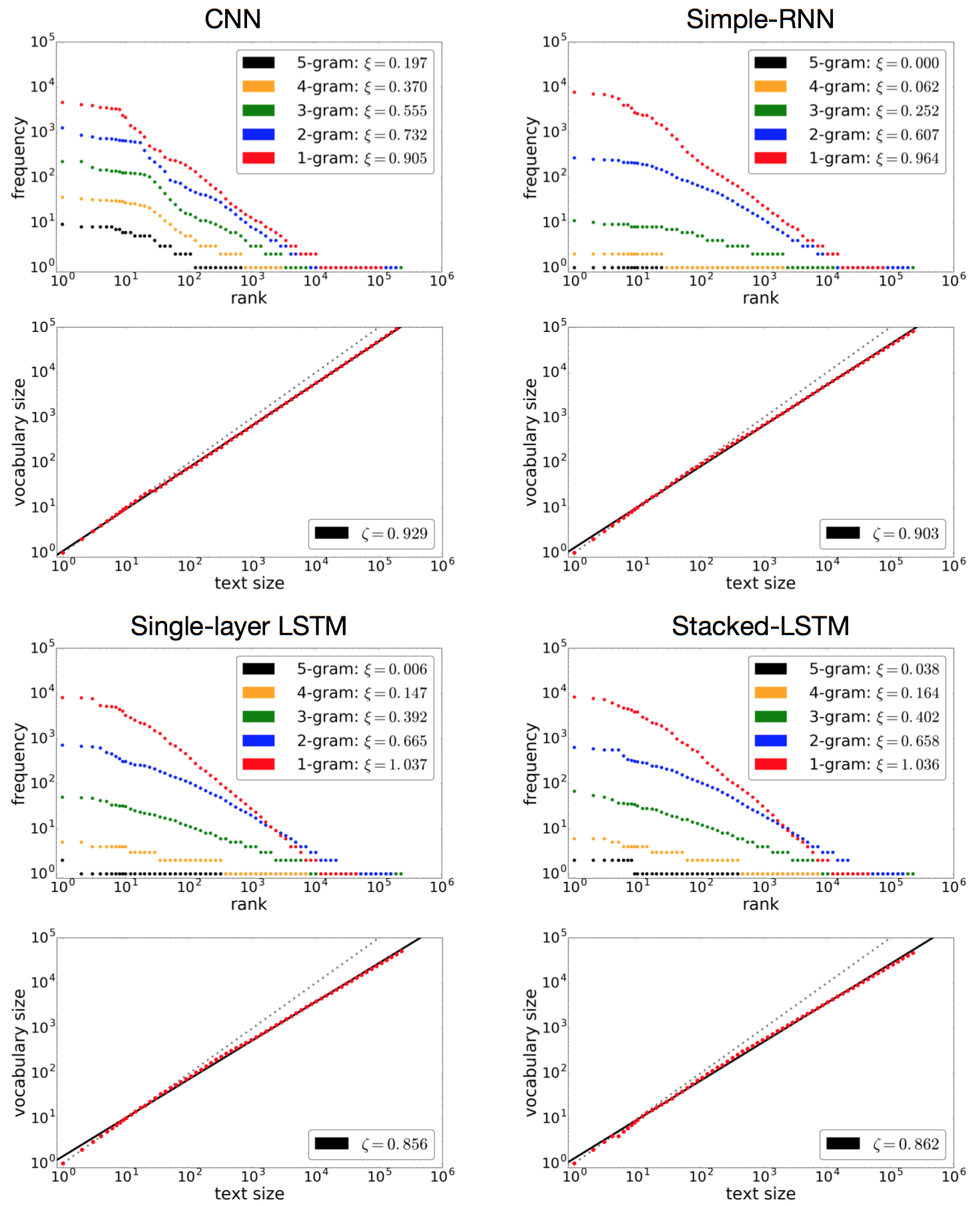}
\end{center}
{{\bf S1Fig.} Comparison of the rank-frequency distribution and vocabulary growth of 
different models for the Complete Works of Shakespeare. Each pair of graphs consists of the 
rank-frequency distribution (upper graph) and the vocabulary growth (lower graph). 
% The 
% results were obtained with a CNN (upper left), simple RNN (upper right), single-layer 
% LSTM (bottom left), and stacked LSTM (bottom right, the same graphs from 
% \figref{shakespeare:zipfheaps}). 
The models had the following specifications. CNN: 8 
layers of one-dimensional convolution with 256 filters having a width of 7 without 
padding and global max pooling after the last convolutional layer. The activation 
function was rectified linear-unit, and batch normalization was applied before activation 
in every convolutional layer. Simple-RNN: 1 layer of RNN with 512 units \add{and an 
output softmax layer.} Single-layer LSTM: 1 layer of LSTM with 512 units \add{and 
an output softmax layer}. Stacked-LSTM: as described in Section 2.}
\label{different_models}
\end{figure}

\begin{figure}[h]
\begin{center}
\includegraphics[width=0.8\textwidth]{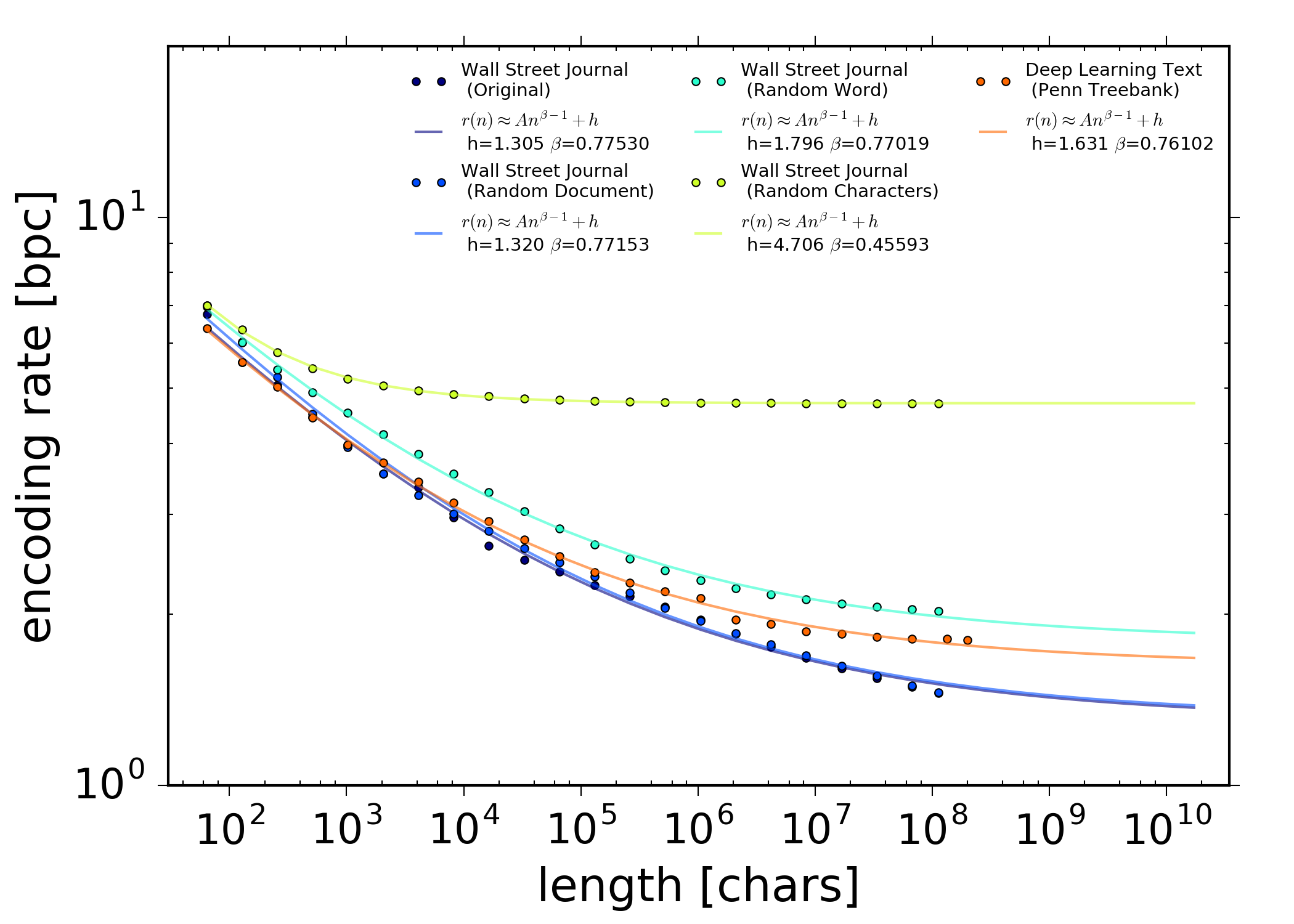}
\end{center}
\add{
{{\bf S2Fig.} Encoding rate decay and fitting functions for the WSJ with various 
shuffling methods and the corresponding pseudo-text generated by the stacked LSTM. 
Let $X_1^n$ be a text of length $n$ characters, and let $R(X_1^n)$ be its size in bits 
after compression. Then the code length per unit, i.e., the encoding rate, is defined as 
$r(n)=R(X_1^n)/n$. The more predictable the text is, the smaller  $r(n)$ becomes;
therefore, $r(n)$ is smaller for longer $n$, exhibiting decay. The fitting function here is 
a power ansatz function, $f(n) = A n^{\beta-1} + h$, proposed by \citep{Hilberg}, and 
the compressor was PPMd, using the 7zip application (refer to \citep{entropy16} for 
details). In addition to the original text, the WSJ was shuffled at the character, word, and 
document levels. The decay of the pseudo-text is situated between the decays of the 
word- and document-shuffled versions, indicating clearly that its predictability is 
situated between the two and suggesting that the pseudo-text has lower predictability as 
compared to the original and document-shuffled WSJ. }}
\label{encoding_rate}
\end{figure}

\begin{figure}[h]
\begin{center}
\includegraphics[width=\textwidth]{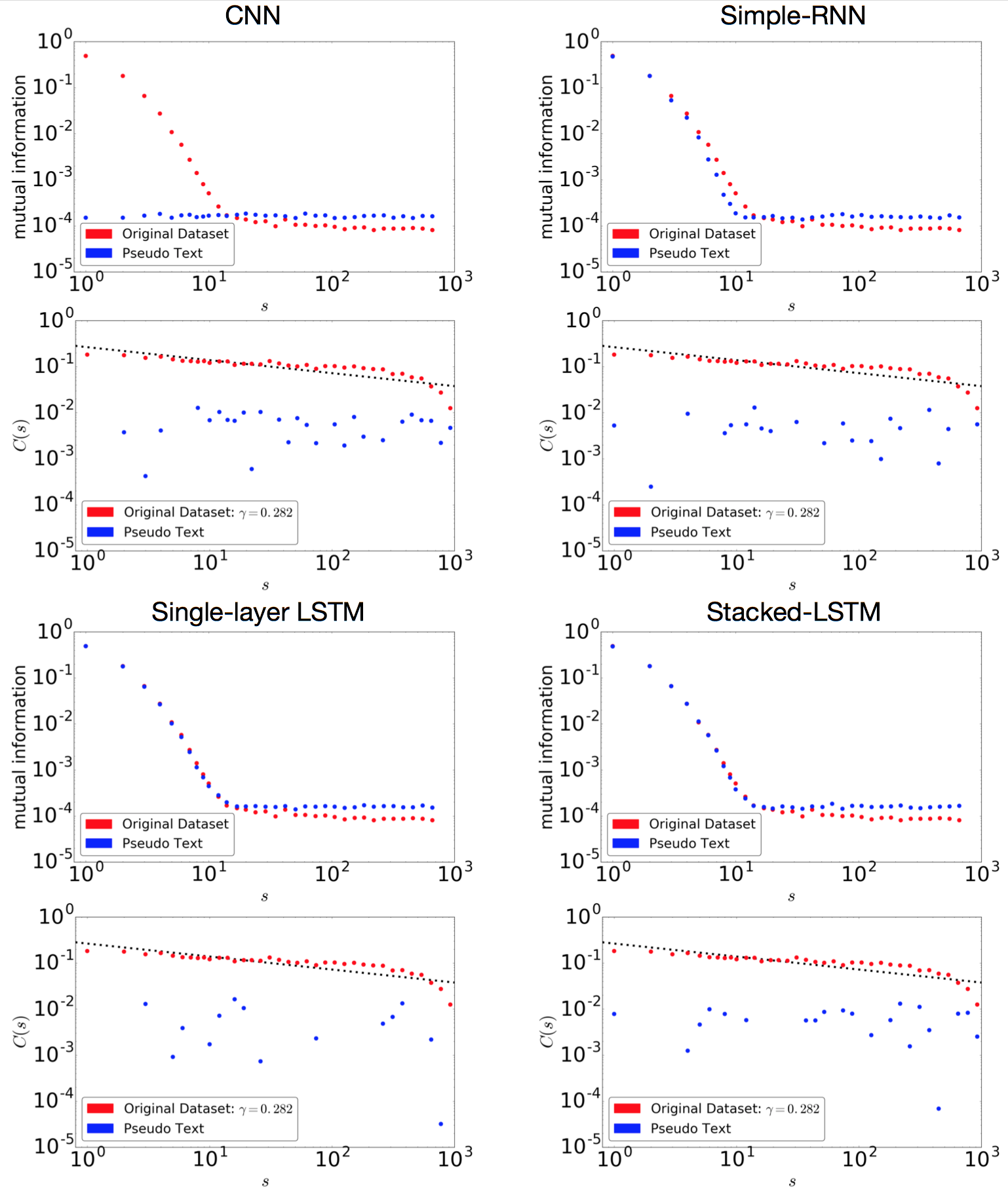}
\end{center}
\add{
{{\bf S3Fig.} Comparison of the mutual information and autocorrelation function for 
pseudo-texts generated by different models on the Complete Works of Shakespeare. Each pair of 
graphs represents the mutual information (upper) and the autocorrelation function 
(lower). The results were obtained with a CNN (upper left), simple RNN (upper right), 
single-layer LSTM (lower left), and stacked LSTM (bottom right, the same graphs from 
\figref{shakespeare:zipfheaps}).
For the specifications of every model, see the caption of S1Fig.
}
}
\label{different_models_s3}
\end{figure}

\begin{figure}[h]
\begin{center}
\includegraphics[width=\textwidth]{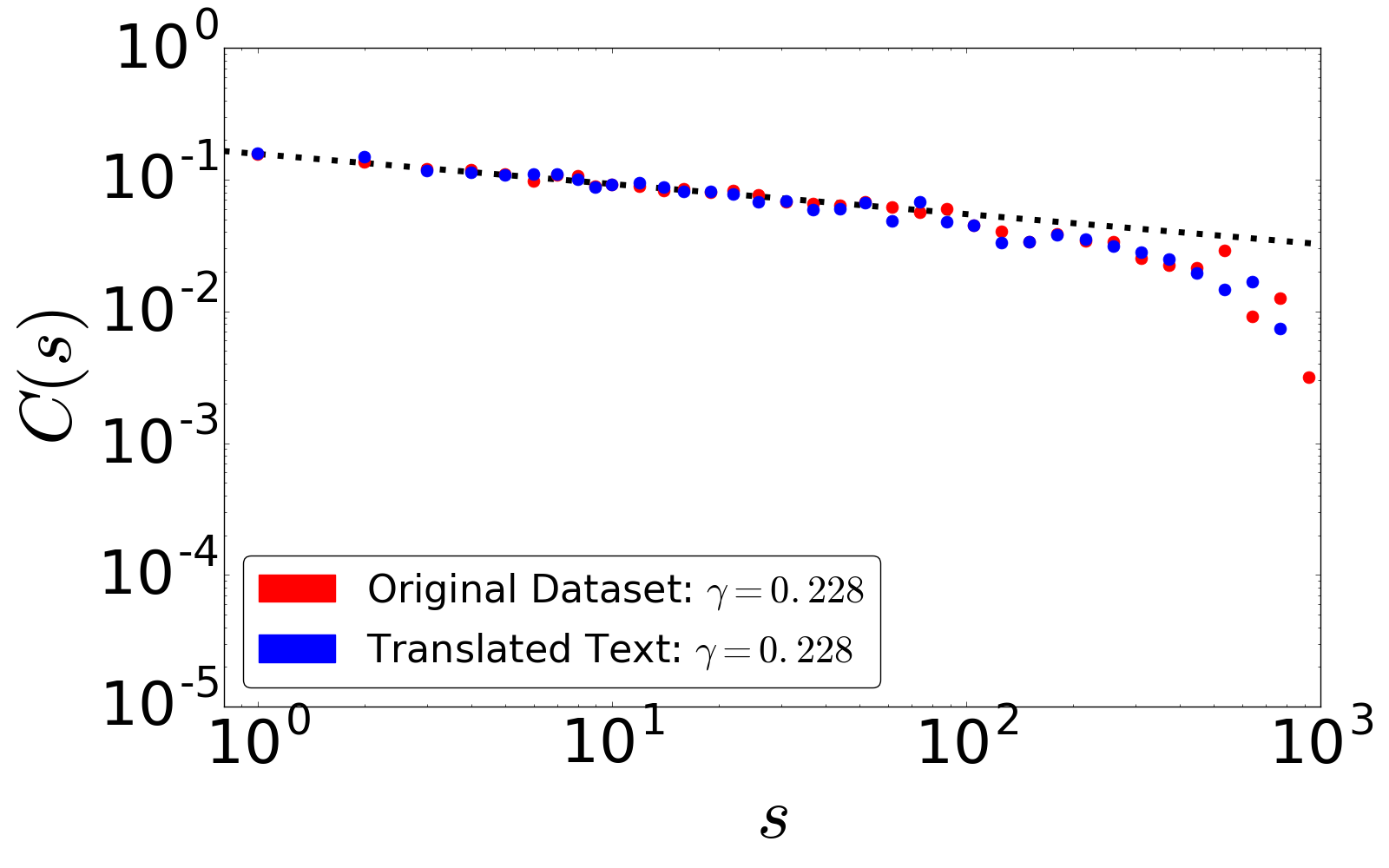}
\end{center}
\add{
{{\bf S4Fig.} Autocorrelation functions of the original text of Les Mis\'erables (V. 
Hugo, 621,641 words) in French and its corresponding text translated into English by 
the Google Cloud Translation API {\tt https://cloud.google.com/translate/}, which is 
based on neural machine translation. Because of the API's requirements, the original 
text was split into 5000 characters to obtain the translated text. Despite the results given 
in \secref{sec:acf}, the translated text exhibits long-range correlation as measured by 
the autocorrelation function. This result does not contradict our observation in 
\secref{sec:acf}, because translation does not radically change the order of words and 
the translation system has the capacity to output rare words.}}
\label{translation}
\end{figure}

\end{document}